# On Bayesian Network Approximation by Edge Deletion

**Arthur Choi** and **Hei Chan** and **Adnan Darwiche**
Computer Science Department
University of California, Los Angeles
Los Angeles, CA 90095
{*aychoi,hei,darwiche*}@cs.ucla.edu

## Abstract

We consider the problem of deleting edges from a Bayesian network for the purpose of simplifying models in probabilistic inference. In particular, we propose a new method for deleting network edges, which is based on the evidence at hand. We provide some interesting bounds on the KL-divergence between original and approximate networks, which highlight the impact of given evidence on the quality of approximation and shed some light on good and bad candidates for edge deletion. We finally demonstrate empirically the promise of the proposed edge deletion technique as a basis for approximate inference.

## 1 INTRODUCTION

Classical algorithms for exact probabilistic inference have a complexity which is parameterized by the network topology [Jensen *et al.*, 1990; Lauritzen and Spiegelhalter, 1988; Zhang and Poole, 1996; Dechter, 1996; Darwiche, 2001]. In particular, it is well known that exact inference can be performed exponential only in the treewidth of a given network, where treewidth is a graph theoretic parameter that measures network connectivity. When the treewidth is high, one may then consider simplifying the network by deleting some edges to reduce its treewidth and then run exact inference on the simplified network. This approach would then lead to a class of approximate inference algorithms, which result from applying exact inference to an approximate network.

Clearly, the quality of approximate inference in this case would depend on the quality of approximate networks one constructs as a result of edge deletion. In previous work on edge deletion, most results focused on the quality of approximation between the prior distributions induced by the original and approximate networks, typically by considering the KL–divergence between the two [Cover and Thomas, 1991]. However, it is known that the KL–divergence may be small between two distributions, yet blow up when conditioning on a particular evidence [Koller, 1996]. Therefore, bounding the KL–divergence would not necessarily provide guarantees on conditional queries.

In this paper, we propose an edge deletion method which is sensitive to the available evidence. The method was motivated by the following observations. First, by deleting an edge $Y \to X$ from a network, we are in essence changing the conditional probability table (CPT) of variable $X$, since $X$ will have one less parent in the new network. Moreover, it is known that if the current evidence $\mathbf{e}$ fixes the value of variable $Y$, then the edge $Y \to X$ can be deleted and the CPT for $X$ can be modified, while yielding a simpler network which corresponds exactly to the original network for any query of the form $\alpha, \mathbf{e}$. The question now is: what if the current evidence does not fix the value of $Y$, but the probability of $Y$ is extreme given evidence $\mathbf{e}$? Can we in this case delete $Y \to X$ and still expect to get a good approximate network? In particular, would the approximate results converge to the exact ones as the posterior of $Y$ given $\mathbf{e}$ converges to an extreme distribution? We will indeed provide a bound and analysis that give some interesting insights on this matter.

We also empirically evaluate the edge deletion methods as a basis for approximate inference, by running exact inference on the approximated network. This method of approximate inference is interesting as it provides a tradeoff between efficiency and quality of approximation, through control over the deleted edges. It is therefore in the same spirit as generalized belief propagation [Yedidia *et al.*, 2000], except that it is independent of the specific method used for exact inference. As we shall see, the formulation based on edge deletion appears to provide new grounds for analysis.

This paper is structured as follows. In Section 2, we define the semantics of edge deletion. In Section 3,

we introduce some interesting bounds on the KL–divergence between the original and approximate networks, for both deleting a single edge, and deleting multiple edges. In Section 4 we address one of the key subtleties relating to our edge deletion method, and in Section 5 we present empirical results on an approximate inference method based on edge deletion. Section 6 discusses previous work in edge deletion, and Section 7 closes with some concluding remarks. Proofs of Theorems are included in the Appendix.

## 2 DELETING EDGES

Deleting an edge $Y \to X$ from a network entails more than simply removing an edge from the graph; one must also have a way of updating the CPT of variable $X$, which has one less parent after deletion. Other notions of edge deletion studied in the past, which we will review in Section 6, motivated their deletion by asserting, in a particular sense, a conditional independence between $Y$ and $X$. However, as these methods did not specifically take evidence into account, an approximation may look good before conditioning on evidence, but could potentially be a bad one afterwards.

Consider a variable $Y$ in a network $N$ and some evidence $\mathbf{e}$ that fixed the value of $Y$. We can then delete each outgoing edge of $Y$, and assume the fixed value of $Y$ in the CPT of each of its children $X$. This method will give a network with exact results for queries of the form $\alpha, \mathbf{e}$. But what if the value of a variable is *almost* determined by evidence $\mathbf{e}$? Then perhaps we can try to weight the CPT for $X$ by the posterior probability distribution of the parent $Y$.

Assume that we have a network $N$ in which node $X$ has parents $Y$ and $\mathbf{U}$, where $\mathbf{U}$ may be empty. We want to approximate this network by another, say $N'$, which results from removing the edge $Y \to X$. This approximation will be done given a piece of evidence $\mathbf{e}$, by replacing the CPT $\Theta_{X|Y\mathbf{U}}$ of variable $X$ in network $N$ by the CPT $\Theta'_{X|\mathbf{U}}$ as defined below.

**Definition 1 (Edge Deletion)** *Let $N$ be a Bayesian network with node $X$ having parents $Y$ and $\mathbf{U}$. The network $N'$ which results from deleting edge $Y \to X$ from $N$ given evidence $\mathbf{e}$ is defined as follows:*

- *$N'$ has the same structure as $N$ except that edge $Y \to X$ is removed.*
- *The CPT for variable $X$ in $N'$ is given by:*

$$\theta'_{x|\mathbf{u}} \stackrel{def}{=} \sum_y \theta_{x|y\mathbf{u}} Pr(y|\mathbf{e}).$$

- *The CPTs for variables other than $X$ in $N'$ are the same as those in $N$.*

Note that this approximation assumes that we have the posteriors $Pr(Y|\mathbf{e})$ in the original network $N$. This probability is typically not available, as the network $N$ must already be difficult, for us to be interested in deleting some of its edges. We will indeed address this point in Section 4, but for now we will pretend that we have this posterior probability. Note also that the probabilities $Pr'(Y|\mathbf{e})$ in the approximate network may not equal the original probabilities $Pr(Y|\mathbf{e})$ when the edge $Y \to X$ is cut as given above.

Intuitively, this edge deletion method is equivalent to creating an auxiliary root node $Y'$ whose CPT $\Theta_{Y'}$ is $Pr(Y|\mathbf{e})$ and then replacing the original parent $Y$ with the new node $Y'$. This can be shown by eliminating variable $Y'$ from the new network, which leads to replacing the CPT of $X$ by the one proposed in Definition 1.

## 3 QUALITY OF APPROXIMATION

We consider in this section the quality of networks generated by the proposed edge deletion method. In particular, we provide bounds on the KL–divergence between the conditional distributions $Pr(.|\mathbf{e})$ and $Pr'(.|\mathbf{e})$ induced by the original and approximate networks $N$ and $N'$, where KL is defined as follows [Cover and Thomas, 1991]:

$$KL(Pr, Pr') \stackrel{def}{=} \sum_w Pr(w) \log \frac{Pr(w)}{Pr'(w)}.$$

### 3.1 DELETING A SINGLE EDGE

Our intuition suggests that deleting an edge out of a variable whose value is *almost* determined given the evidence may yield a reasonable approximation. The following bound lends some support to that intuition:

**Theorem 1** *Let $N$ and $N'$ be two Bayesian networks as given in Definition 1. We then have:*

$$KL(Pr(.|\mathbf{e}), Pr'(.|\mathbf{e})) \leq \log \frac{Pr'(\mathbf{e})}{Pr(\mathbf{e})} + ENT(Y|\mathbf{e}),$$

*where $ENT(Y|\mathbf{e})$ is the entropy of $Y$ given $\mathbf{e}$, and is defined as follows:*

$$ENT(Y|\mathbf{e}) \stackrel{def}{=} -\sum_y Pr(y|\mathbf{e}) \log Pr(y|\mathbf{e}).$$

Note that the $ENT(Y|\mathbf{e}) = 0$ if and only if $Pr(y|\mathbf{e}) = 1$ for some $y$; that is, the entropy of $Y$ given $\mathbf{e}$ is zero if and only if the value of $Y$ is determined. Now consider the following condition in which our bound becomes an equality:

**Theorem 2** Let $N$ and $N'$ be two Bayesian networks as given in Definition 1. If the CPT for variable $X$ is deterministic:
$$\theta_{x|y\mathbf{u}} \in \{0,1\} \quad (1)$$
and
$$\theta_{x|y\mathbf{u}} = \theta_{x|y'\mathbf{u}} = 1 \text{ only if } y = y' \quad (2)$$
then
$$KL(Pr(.|\mathbf{e}), Pr'(.|\mathbf{e})) = \log \frac{Pr'(\mathbf{e})}{Pr(\mathbf{e})} + ENT(Y|\mathbf{e}).$$

Note that Condition 2 is satisfied when the CPT for $X$ has no context–specific independence. Moreover, both conditions are satisfied if $X$ is a parity function of its parents, as one typically finds, for example, in networks for error-correcting codes [Frey and MacKay, 1997].

The more certain we are of the value of $Y$ given $\mathbf{e}$, the more extreme $Pr(Y|\mathbf{e})$ is, and the lower $ENT(Y|\mathbf{e})$ is. This suggests that if the value of $Y$ is *almost* determined, then the $ENT(Y|\mathbf{e})$ term in the bound on the KL–divergence is negligible, and we may get a good approximation. Moreover, the $\log Pr'(\mathbf{e})/Pr(\mathbf{e})$ term may be negative, and the entropy term need not be small for us to get a good approximation. However, as we demonstrate in Appendix B, there are particular situations where $ENT(Y|\mathbf{e})$ can be arbitrarily close to zero, but where $\log Pr'(\mathbf{e})/Pr(\mathbf{e})$ can be unbounded.

### 3.2 DELETING MULTIPLE EDGES

We can extend the bound given in Theorem 1 about a single edge deletion, to a bound on multiple deletions, where the entropy term is additive:

**Theorem 3** Let $N'$ be a network obtained from network $N$ by deleting multiple edges $Y \to X$ as given by Definition 1, deleting at most one incoming edge per node $X$. Then
$$KL(Pr(.|\mathbf{e}), Pr'(.|\mathbf{e})) \leq \log \frac{Pr'(\mathbf{e})}{Pr(\mathbf{e})} + \sum_{Y \to X} ENT(Y|\mathbf{e}).$$

Moreover, if the CPTs for each $X$ satisfy Conditions 1 and 2, then
$$KL(Pr(.|\mathbf{e}), Pr'(.|\mathbf{e})) = \log \frac{Pr'(\mathbf{e})}{Pr(\mathbf{e})} + \sum_{Y \to X} ENT(Y|\mathbf{e}).$$

### 4 FIXED POINTS

Deleting edges $Y \to X$ by Definition 1 assumes that we know the distribution on $Y$ given $\mathbf{e}$. If we are interested in approximating our network $N$, then computing $Pr(Y|\mathbf{e})$ is itself likely to be difficult computationally. In the approximated network, however, computing the posteriors of $Y$ is likely to be easy. We will therefore use the approximate network to approximate $Pr(Y|\mathbf{e})$ using an iterative method as discussed below.

Suppose that $\widehat{N}$ is the network that we obtain by deleting edges. We first assume that each $\widehat{Pr}_{t=0}(Y|\mathbf{e})$ is uniform, and then cut edges $Y \to X$ by setting the CPT for variable $X$ in $\widehat{N}$ as follows:
$$\theta'_{x|\mathbf{u}} = \sum_y \theta_{x|y\mathbf{u}} \widehat{Pr}_{t=0}(y|\mathbf{e}).$$

We then apply exact inference to the approximate network. If $\widehat{Pr}_t(Y|\mathbf{e})$ are the posteriors of $Y$ at iteration $t$ in the approximate network, then the CPT for $X$ is updated as follows:
$$\theta'_{x|\mathbf{u}} = \sum_y \theta_{x|y\mathbf{u}} \widehat{Pr}_t(y|\mathbf{e}).$$

We repeat this process until we find that for all edges that we delete, all $\widehat{Pr}_t(Y|\mathbf{e}) = \widehat{Pr}_{t+1}(Y|\mathbf{e})$, or that they are within some threshold $\epsilon$ from one iteration to the next. At this point, we say that we have converged, and that $\widehat{Pr}(Y|\mathbf{e})$ are a fixed point for $\widehat{N}$.

We will compare in the following section the quality of our approximations when computed based on the true posteriors $Pr(Y|\mathbf{e})$ and the one obtained by the above iterative method, showing that the fixed point method tends to work quite well on the given benchmarks.

### 5 EMPIRICAL ANALYSIS

We discuss here experimental results on edge deletion. In particular, we compare iterative belief propagation [Pearl, 1988; Murphy *et al.*, 1999] as an approximate inference algorithm, with exact inference on a network with deleted edges. Moreover, we compare the deletion of edges $Y \to X$ based on the true probabilites $Pr(Y|\mathbf{e})$ and the fixed point probabilites $\widehat{Pr}(Y|\mathbf{e})$.

We use the jointree algorithm to perform exact inference on the network with deleted edges. Moreover, we make the following choices in our experiments.

*Cost of computation:* We use the largest cluster size in the jointree to measure the difficulty of exact inference. For belief propagation, we assume that difficulty depends on the number of loops to observe convergence, where the complexity of each loop depends on the number and sizes of the network CPT's, and in particular, the size of the largest CPT.

*Deleted edges:* For each network we consider, we cut different sets of edges to control the size of the largest cluster in the corresponding jointree. We do this by fixing a variable order that induces a jointree.[1] We then delete edges to find an approximated network which has a jointree whose largest cluster is smaller than given thresholds, using the same variable ordering. We decide on edges to delete based on a heuristic based on reducing bucket sizes to satisfy a given threshold in a bucket elimination procedure [Dechter, 1996].

---
[1] We use min-fill/min-size heuristics, or by using orders found otherwise. For example, orders are available for networks in Aalborg DSS's repository.

*Evidence:* For each of a given number of trials (typically 50 to 200 trials), we set evidence on all leaves by sampling based on the prior probabilities of each individual variable, to approximately simulate evidence that we may likely observe.

*Convergence:* For each piece of evidence, we approximate inference by belief propagation and by edge deletion for a given set of thresholds on the largest cluster size. When appropriate, we decide on convergence when the marginal of every variable of an iteration is within $10^{-8}$ of the previous iteration, trying at most 100 iterations. If we do not observe convergence within this many iterations, we evaluate the instance based on the state of the network in its last iteration.

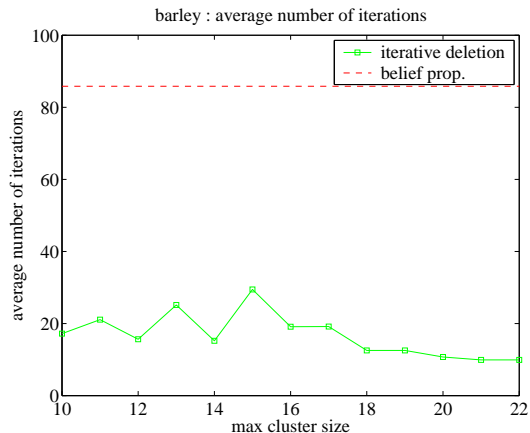

Figure 1: Loops until convergence in barley network

First consider the barley network, which has a jointree with a normalized maximum cluster size of about 22.79, which is the $\log_2$ of the number of entries in the largest table of a cluster. The barley network also has a large CPT containing $40,320$ entries. Although the largest cluster in the jointree still has 180 times more entries then the largest CPT in the network, if belief propagation takes many iterations to converge, we get an approximation of posterior marginals, but with less attractive benefits in time savings.

Consider Figure 1, which compares the average number of loops required until convergence. In 70 trials, belief propagation converges in less than half the instances when given 100 iterations to converge, taking 86 iterations on average overall. In comparison, the iterative version of edge deletion always converged within 100 iterations for all trials, and took 16 iterations on average for this network. When we use a threshold on the maximum cluster size of $\log_2 40,320 = 15.30$ or smaller, the largest tables computed during belief propagation and edge deletion are of comparable sizes, but edge deletion still needs only 21 iterations on average to converge.

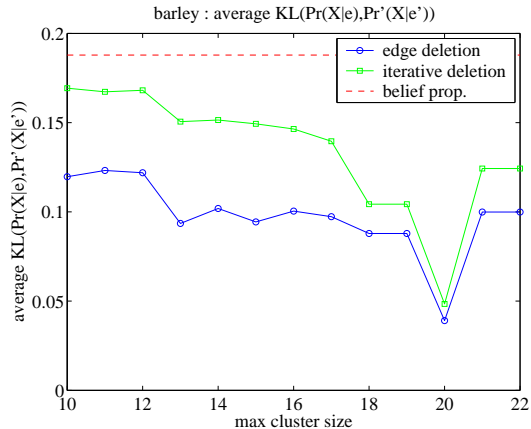

Figure 2: Quality of approximation in barley network

We judge the quality of approximation with two measures. One measure is in terms of the average number of flips observed in a trial. We say that a flip occurs when the most likely state of an individual variable with respect to the original posterior distribution is no longer the most likely state in the approximated one. The other measure is in terms of the average KL–divergence between the true and approximated conditional probabilities of non-evidence variables. Note that in our figures, curves are not always smooth, since different thresholds yield different sets of edges deleted, and smaller sets of edges deleted do not necessarily yield more accurate approximations.

Consider first Figure 2, which compares the quality of the approximation given by edge deletion and belief propagation in the barley network, based on the KL–divergence. We see in this case, both methods of edge deletion compare favorably to belief propagation for all given thresholds on the cluster size. At a threshold of 22, we find that the largest cluster size of the approximation is 20.47, and so the number of entries in that cluster table is 20% the size of that in the original network. With a threshold of 20, the largest table is 3.54% the size; with a threshold of 10, it is $7.72 \cdot 10^{-3}\%$ the size.

Figures 3 and 4 compare the quality of approximations in the munin1 and munin3 networks. We observe that the quality of approximation tends to degrade with a stricter threshold on the largest cluster size. Note that we are deciding on which edges to delete without regard to the particular instantiation of evidence, and we see that our approximation schemes compare favorably to belief propagation to a point, with varying degrees of computational savings. In munin3, both methods of edge deletion compare favorably against belief propagation, in both measures, up until a threshold of 11. The largest normalized cluster size in the original network is 17.26 compared to 10.97 in the approximated network at that threshold. Thus, the largest table in the jointree for the approximated network is 1.28% of that of the original network.

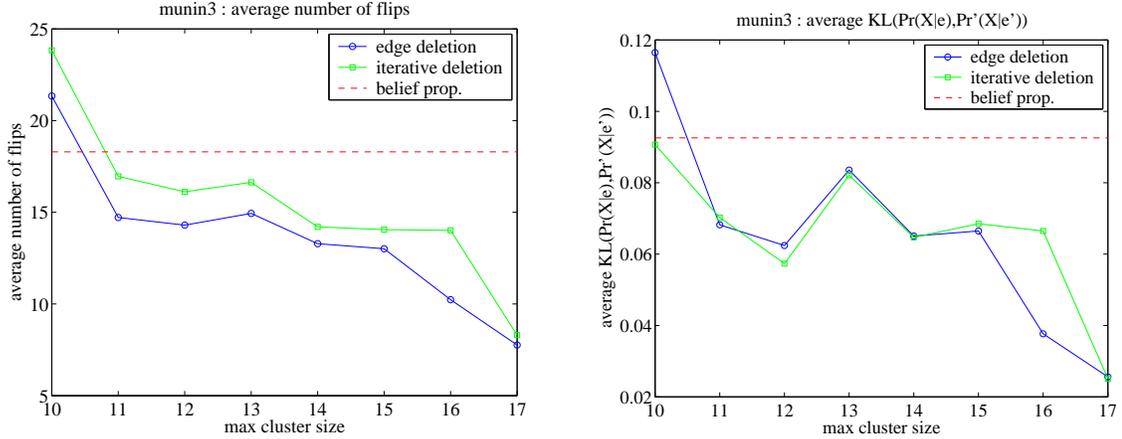

Figure 3: Quality of approximation in munin3 network.

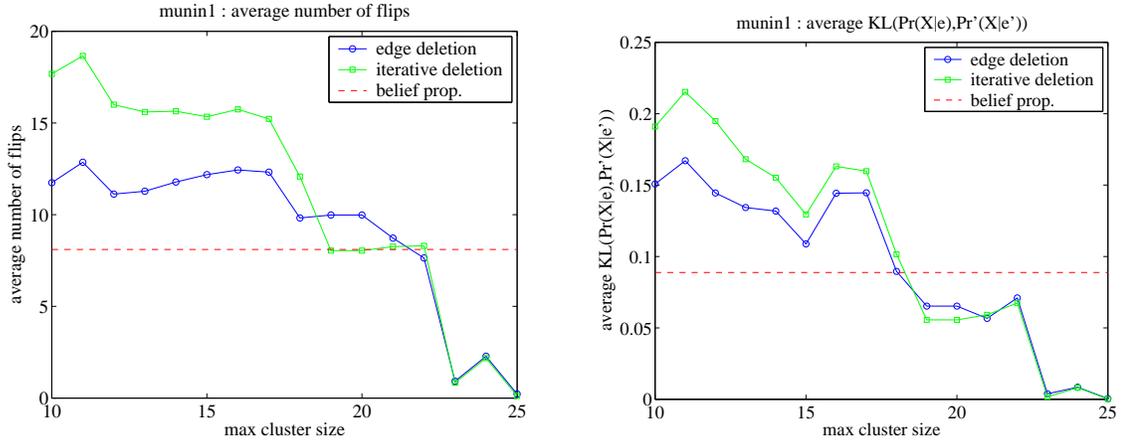

Figure 4: Quality of approximation in munin1 network.

In most of our experiments, we tend to observe that the iterative version of edge deletion is very close to the version that uses the true probabilities $Pr(Y|e)$ for larger thresholds, and typically fewer edges deleted, and become less comparable for smaller and smaller thresholds. In munin3, they perform similarly over most of the given thresholds. In munin1, we see that both versions of edge deletion are comparable up to a point, where the iterative method tends to fare worse for smaller thresholds.

Figure 5 provides a summary of the comparisons between our methods of approximation. In these tables, we measure how belief propagation performs in terms of the average percentage of flips observed, and by average $KL(Pr(X|e), Pr'(X|e))$ over all non-evidence variables. We then find the threshold on the maximum cluster size in which edge deletion using true probabilities $Pr(Y|e)$ compares favorably to belief propagation, and report the savings in cluster size with respect to the cluster size of the original network. We also report the quality of approximation for the iterative version of edge deletion that uses fixed point probabilities $\widehat{Pr}(Y|e)$, for the same threshold. We find that both versions of edge deletion compare well with belief propagation, and can do so with substantial degrees of computational savings in terms of reduction of the largest cluster size. Further, we see that when deletion with true probabilities $Pr(Y|e)$ does well, deletion

| network | cluster % size | average % flips | | |
|---------|---------|------|------|------|
| | | BP | ED | ID |
| munin1 | 7.31% | 5.19% | 4.90% | 5.32% |
| munin2 | 16.65% | 0.56% | 0.39% | 0.35% |
| munin3 | 1.28% | 2.14% | 1.72% | 1.98% |
| munin4 | 37.55% | 1.47% | 0.87% | 0.82% |
| barley | 3.52% | 24.28% | 17.21% | 19.89% |
| pigs | 11.08% | 1.86% | 1.11% | 1.00% |

| network | cluster % size | average KL | | |
|---------|---------|------|------|------|
| | | BP | ED | ID |
| munin1 | 1.17% | 0.0888 | 0.0652 | 0.0556 |
| munin2 | 16.65% | 0.0134 | 0.0116 | 0.0112 |
| munin3 | 1.28% | 0.0926 | 0.0682 | 0.0702 |
| munin4 | 37.55% | 0.0416 | 0.0261 | 0.0247 |
| barley | 0.01% | 0.1879 | 0.1197 | 0.1693 |
| pigs | 11.08% | 0.0042 | 0.0020 | 0.0020 |

Figure 5: Savings in maximum cluster size (% size) at the point where edge deletion (ED) outperforms belief propagation (BP), in terms of flips and average KL–divergence. KL for iterative version of edge deletion (ID) also shown. In barley, ED compares favorably in terms of the KL to BP for all thresholds down to our limit of 10.

with fixed point probabilties also tends to do well.

## 6  PREVIOUS WORK

Prior work in approximating Bayesian networks by edge deletion focused primarily on the effects of deletion on the prior distribution. In [Kjærulff, 1994], edges are deleted, not in the model itself, but on the moralized independence graph induced by the network, and its simplifications are specific to jointree based algorithms. Weak dependencies are sought between pairs of variables within particular cliques of the jointree, and a conditional independence is asserted based on the variables that appear in the clique. The author showed that the KL–divergence between the original distribution and the one where a conditional independence is asserted is equal to the conditional mutual information [Cover and Thomas, 1991] of the variables involved. He further showed that the KL–divergence is additive, and that for each conditional independence asserted, the divergence is computable locally. It is possible to recover a model from the approximated jointree, but the structure may not be unique, and the parameterization for it may not be easily determined. The divergence is for prior distributions, and a form for the posterior distribution was cited as future work. More recently, [Paskin, 2003] effectively employed this simplification in a Gaussian graphical model for simultaneous localization and mapping for mobile robotics.

In [van Engelen, 1997], edges are deleted in the model, as we did here. For a node $X$ with parents $Y\mathbf{U}$, we can cut edge $Y \to X$ by replacing each $\theta_{x|y\mathbf{u}}$ with $\theta'_{x|\mathbf{u}} = Pr(x|\mathbf{u})$; essentially, we are asserting that $X$ and $Y$ are conditionally independent given $\mathbf{U}$. The KL–divergence between the prior distributions is again the conditional mutual information, and is additive if we delete at most a single incoming edge per node. However, to make this approximation, one must be able to compute the quantities $Pr(x|\mathbf{u})$. Although these are local quantities, they may require global computations. Further, approximations may not be good after conditioning on unlikely evidence.

## 7  CONCLUSION

We proposed a method for deleting edges from a Bayesian network, which is sensitive to the evidence at hand. We provided some bounds on the KL–divergence between the original and approximated network, given evidence. The bounds shed light on when this method is expected to provide good approximations. We also evaluated empirically an approximate inference algorithm which is based on deleting edges against belief propagation, and showed that the edge deletion method holds good promise. The method we used to decide on which edges to delete is a bit primitive, as it is based on a fixed variable order and does not exploit the given evidence for making its choices. We are currently working on a more sophisticated scheme for this purpose, which may significantly improve the quality of approximations obtained by the proposed method of edge deletion.

**Acknowledgments**

This work has been partially supported by Air Force grant FA9550-05-1-0075 and MURI grant N00014-00-1-0617.

**References**

[Cover and Thomas, 1991] Thomas M. Cover and Joy A. Thomas. *Elements of information theory*. Wiley-Interscience, 1991.

[Darwiche, 2001] Adnan Darwiche. Recursive conditioning. *Artificial Intelligence*, 126(1-2):5–41, 2001.

[Dechter, 1996] Rina Dechter. Bucket elimination: A unifying framework for probabilistic inference. In *UAI*, pages 211–219, 1996.

[Jensen *et al.*, 1990] F. V. Jensen, S.L. Lauritzen, and K.G. Olesen. Bayesian updating in recursive graphical models by local computation. *Computational Statistics Quarterly*, 4:269–282, 1990.

[Frey and MacKay, 1997] Brendan J. Frey and David J. C. MacKay. A revolution: Belief propagation in graphs with cycles. In *NIPS*, pages 479–485, 1997.

[Kjærulff, 1994] Uffe Kjærulff. Reduction of computational complexity in Bayesian networks through removal of weak dependences. In *UAI*, pages 374–382, 1994.

[Koller, 1996] Daphne Koller. Evidence-directed belief network simplification. In *Working Notes of the AAAI Fall Symposium on Flexible Computation in Intelligent Systems*, 1996.

[Lauritzen and Spiegelhalter, 1988] S. L. Lauritzen and D. J. Spiegelhalter. Local computations with probabilities on graphical structures and their application to expert systems. *Journal of Royal Statistics Society, Series B*, 50(2):157–224, 1988.

[Murphy *et al.*, 1999] Kevin P. Murphy, Yair Weiss, and Michael I. Jordan. Loopy belief propagation for approximate inference: An empirical study. In *UAI*, pages 467–475, 1999.

[Paskin, 2003] Mark A. Paskin. Thin junction tree filters for simultaneous localization and mapping. In *IJCAI*, pages 1157–1166, 2003.

[Pearl, 1988] Judea Pearl. *Probabilistic Reasoning in Intelligent Systems: Networks of Plausible Inference*. Morgan Kaufmann, 1988.

[van Engelen, 1997] Robert A. van Engelen. Approximating Bayesian belief networks by arc removal. *PAMI*, 19(8):916–920, 1997.

[Yedidia *et al.*, 2000] Jonathan S. Yedidia, William T. Freeman, and Yair Weiss. Generalized belief propagation. In *NIPS*, pages 689–695, 2000.

[Zhang and Poole, 1996] Nevin Lianwen Zhang and David Poole. Exploiting causal independence in bayesian network inference. *JAIR*, 5:301–328, 1996.

## A  PROOFS

Say we want to compare network $N$ with a network $N'$ by deleting edges in $N$; we can analyze and compare the two as if they had the same structure. For example, say node $X$ has parents $Y\mathbf{U}$; if we wanted to delete $Y \to X$ in $N$, then a network $N'$ would have a CPT $\Theta'_{X|Y\mathbf{U}}$ instead of $\Theta'_{X|\mathbf{U}}$ where for all $y$, $\theta'_{x|y\mathbf{u}} = \theta'_{x|\mathbf{u}}$. For the sake of analysis, we shall assume that we "delete" edges in this manner from here on.

**Theorem 4** *Let $N$ and $N'$ be two Bayesian networks that have the same structure and that agree on all CPTs except the one for variable $X$, with parents $Y$ and $\mathbf{U}$. Then with evidence $\mathbf{e}$, we have*

$$KL(Pr(.|\mathbf{e}), Pr'(.|\mathbf{e}))$$
$$= \log \frac{Pr'(\mathbf{e})}{Pr(\mathbf{e})} -$$
$$\sum_{y\mathbf{u}} Pr(y\mathbf{u}|\mathbf{e}) \sum_x Pr(x|y\mathbf{ue}) \log \frac{\theta'_{x|y\mathbf{u}}}{\theta_{x|y\mathbf{u}}}.$$

*Further, if the CPT of $X$ is deterministic, we have*

$$KL(Pr(.|\mathbf{e}), Pr'(.|\mathbf{e}))$$
$$= \log \frac{Pr'(\mathbf{e})}{Pr(\mathbf{e})} + \sum_{y\mathbf{u}} Pr(y\mathbf{u}|\mathbf{e}) KL(\Theta_{X|y\mathbf{u}}, \Theta'_{X|y\mathbf{u}}).$$

**Proof of Theorem 4.**

$$KL(Pr(.|\mathbf{e}), Pr'(.|\mathbf{e}))$$
$$= \sum_w Pr(w|\mathbf{e}) \log \frac{Pr(w|\mathbf{e})}{Pr'(w|\mathbf{e})}$$
$$= \sum_w Pr(w|\mathbf{e}) \log \frac{Pr'(\mathbf{e})}{Pr(\mathbf{e})} -$$
$$\sum_{w \models \mathbf{e}} Pr(w|\mathbf{e}) \log \frac{Pr'(w)}{Pr(w)}$$
$$= \log \frac{Pr'(\mathbf{e})}{Pr(\mathbf{e})} - \sum_{xy\mathbf{u}} \sum_{w \models xy\mathbf{ue}} Pr(w|\mathbf{e}) \log \frac{Pr'(w)}{Pr(w)}$$
$$= \log \frac{Pr'(\mathbf{e})}{Pr(\mathbf{e})} - \sum_{xy\mathbf{u}} Pr(xy\mathbf{u}|\mathbf{e}) \log \frac{\theta'_{x|y\mathbf{u}}}{\theta_{x|y\mathbf{u}}}$$
$$= \log \frac{Pr'(\mathbf{e})}{Pr(\mathbf{e})} -$$
$$\sum_{y\mathbf{u}} Pr(y\mathbf{u}|\mathbf{e}) \sum_x Pr(x|y\mathbf{ue}) \log \frac{\theta'_{x|y\mathbf{u}}}{\theta_{x|y\mathbf{u}}}.$$

If the CPT of $X$ is deterministic, then $Pr(x|y\mathbf{ue}) = Pr(x|y\mathbf{u}) = \theta_{x|y\mathbf{u}}$, and we get the latter equality.  △

**Proof of Theorem 1.** Since

$$\log \frac{\theta'_{x|y\mathbf{u}}}{\theta_{x|y\mathbf{u}}} = \log \frac{\sum_{y'} \theta_{x|y'\mathbf{u}} Pr(y'|\mathbf{e})}{\theta_{x|y\mathbf{u}}}$$
$$= \log \Big( Pr(y|\mathbf{e}) + \sum_{y' \neq y} \frac{\theta_{x|y'\mathbf{u}}}{\theta_{x|y\mathbf{u}}} Pr(y'|\mathbf{e}) \Big)$$
$$\geq \log Pr(y|\mathbf{e}),$$

then from the proof of Theorem 4, we have

$$KL(Pr(.|\mathbf{e}), Pr'(.|\mathbf{e}))$$
$$= \log \frac{Pr'(\mathbf{e})}{Pr(\mathbf{e})} - \sum_{xy\mathbf{u}} Pr(xy\mathbf{u}|\mathbf{e}) \log \frac{\theta'_{x|y\mathbf{u}}}{\theta_{x|y\mathbf{u}}}$$

$$\leq \log \frac{Pr'(\mathbf{e})}{Pr(\mathbf{e})} - \sum_{xy\mathbf{u}} Pr(xy\mathbf{u}|\mathbf{e}) \log Pr(y|\mathbf{e})$$

$$= \log \frac{Pr'(\mathbf{e})}{Pr(\mathbf{e})} - \sum_{y} Pr(y|\mathbf{e}) \log Pr(y|\mathbf{e})$$

$$= \log \frac{Pr'(\mathbf{e})}{Pr(\mathbf{e})} + ENT(Y|\mathbf{e}). \triangle$$

**Proof of Theorem 2.** Suppose that $\theta_{x|y\mathbf{u}} \in \{0,1\}$. Then for each $y\mathbf{u}$, there is a unique $x$, call it $x_{y\mathbf{u}}$, for which $\theta_{x_{y\mathbf{u}}|y\mathbf{u}} = 1$. Note that $Pr(x_{y\mathbf{u}}|y\mathbf{u}\mathbf{e}) = Pr(x_{y\mathbf{u}}|y\mathbf{u}) = \theta_{x_{y\mathbf{u}}|y\mathbf{u}} = 1$ in this case. Then, for a given $y\mathbf{u}$, we have

$$\sum_{x} Pr(x|y\mathbf{u}\mathbf{e}) \log \frac{\theta'_{x|y\mathbf{u}}}{\theta_{x|y\mathbf{u}}}$$

$$= Pr(x_{y\mathbf{u}}|y\mathbf{u}\mathbf{e}) \log \frac{\theta'_{x_{y\mathbf{u}}|y\mathbf{u}}}{\theta_{x_{y\mathbf{u}}|y\mathbf{u}}}$$

$$= \log \Big( Pr(y|\mathbf{e}) + \sum_{y' \neq y} \theta_{x_{y\mathbf{u}}|y'\mathbf{u}} Pr(y'|\mathbf{e}) \Big).$$

Suppose now that $\theta_{x|y\mathbf{u}} = \theta_{x|y'\mathbf{u}} = 1$ only if $y = y'$. Since $\theta_{x_{y\mathbf{u}}|y\mathbf{u}} = 1$, we must have $\theta_{x_{y\mathbf{u}}|y'\mathbf{u}} = 0$. Hence,

$$\sum_{x} Pr(x|y\mathbf{u}\mathbf{e}) \log \frac{\theta'_{x|y\mathbf{u}}}{\theta_{x|y\mathbf{u}}} = \log Pr(y|\mathbf{e})$$

Thus we have,

$$KL(Pr(.|\mathbf{e}), Pr'(.|\mathbf{e}))$$

$$= \log \frac{Pr'(\mathbf{e})}{Pr(\mathbf{e})} -$$

$$\sum_{y\mathbf{u}} Pr(y\mathbf{u}|\mathbf{e}) \sum_{x} Pr(x|y\mathbf{u}\mathbf{e}) \log \frac{\theta'_{x|y\mathbf{u}}}{\theta_{x|y\mathbf{u}}}$$

$$= \log \frac{Pr'(\mathbf{e})}{Pr(\mathbf{e})} - \sum_{y\mathbf{u}} Pr(y\mathbf{u}|\mathbf{e}) \log Pr(y|\mathbf{e})$$

$$= \log \frac{Pr'(\mathbf{e})}{Pr(\mathbf{e})} - \sum_{y} Pr(y|\mathbf{e}) \log Pr(y|\mathbf{e})$$

$$= \log \frac{Pr'(\mathbf{e})}{Pr(\mathbf{e})} + ENT(Y|\mathbf{e}). \triangle$$

**Proof of Theorem 3.** First we generalize Theorem 4 to multiple changes:

$$KL(Pr(.|\mathbf{e}), Pr'(.|\mathbf{e}))$$

$$= \log \frac{Pr'(\mathbf{e})}{Pr(\mathbf{e})} - \sum_{w \models \mathbf{e}} Pr(w|\mathbf{e}) \log \frac{Pr'(w)}{Pr(w)}$$

$$= \log \frac{Pr'(\mathbf{e})}{Pr(\mathbf{e})} - \sum_{w \models \mathbf{e}} Pr(w|\mathbf{e}) \sum_{\theta_{x|y\mathbf{u}}, xy\mathbf{u} \sim w} \log \frac{\theta'_{x|y\mathbf{u}}}{\theta_{x|y\mathbf{u}}}$$

$$= \log \frac{Pr'(\mathbf{e})}{Pr(\mathbf{e})} -$$

$$\sum_{\Theta_{X|Y\mathbf{U}}} \sum_{xy\mathbf{u}} \sum_{w \models xy\mathbf{u}\mathbf{e}} Pr(w|\mathbf{e}) \log \frac{\theta'_{x|y\mathbf{u}}}{\theta_{x|y\mathbf{u}}}$$

$$= \log \frac{Pr'(\mathbf{e})}{Pr(\mathbf{e})} - \sum_{\Theta_{X|Y\mathbf{U}}} \sum_{xy\mathbf{u}} Pr(xy\mathbf{u}|\mathbf{e}) \log \frac{\theta'_{x|y\mathbf{u}}}{\theta_{x|y\mathbf{u}}}.$$

Only the CPT's for variables $X$ for the edges $Y \to X$ we delete are different. We delete at most one edge incoming per node, so each edge deleted corresponds to a unique CPT, and we have:

$$KL(Pr(.|\mathbf{e}), Pr'(.|\mathbf{e}))$$

$$= \log \frac{Pr'(\mathbf{e})}{Pr(\mathbf{e})} - \sum_{Y \to X} \sum_{xy\mathbf{u}} Pr(xy\mathbf{u}|\mathbf{e}) \log \frac{\theta'_{x|y\mathbf{u}}}{\theta_{x|y\mathbf{u}}}$$

$$\leq \log \frac{Pr'(\mathbf{e})}{Pr(\mathbf{e})} - \sum_{Y \to X} \sum_{xy\mathbf{u}} Pr(xy\mathbf{u}|\mathbf{e}) \log Pr(y|\mathbf{e})$$

$$= \log \frac{Pr'(\mathbf{e})}{Pr(\mathbf{e})} - \sum_{Y \to X} \sum_{y} Pr(y|\mathbf{e}) \log Pr(y|\mathbf{e})$$

$$= \log \frac{Pr'(\mathbf{e})}{Pr(\mathbf{e})} + \sum_{Y \to X} ENT(Y|\mathbf{e}).$$

The equality relation can be obtained similar to the proof of Theorem 2.$\triangle$

## B  EXAMPLE

Say network $N$ has binary variables $X, Y, Z$, whose structure is implicit in the parametrization where $\theta_{x|y} = \theta_{\bar{x}|\bar{y}} = 1$, $\theta_{z|xy} = \theta_{z|\bar{x}\bar{y}}$ and $\theta_{z|x\bar{y}} = \theta_{z|\bar{x}y} = 1$. We get the following joint probability distributions induced by $N$ and by $N'$ from deleting edge $Y \to X$ in $N$ both conditioned on evidence $Z = z$:

| $w$ | $\Theta_{X|Y}$ | $\Theta'_X$ | $\Theta_Y$ | $\Theta_{Z|XY}$ | $Pr(w)$ | $Pr'(w)$ |
|---|---|---|---|---|---|---|
| $xyz$ | 1 | $\theta_y$ | $\theta_y$ | $\theta_{z|xy}$ | $\theta_y \theta_{z|xy}$ | $\theta_y \theta_y \theta_{z|xy}$ |
| $x\bar{y}z$ | 0 | $\theta_y$ | $\theta_{\bar{y}}$ | 1 | 0 | $\theta_y \theta_{\bar{y}}$ |
| $\bar{x}yz$ | 0 | $\theta_{\bar{y}}$ | $\theta_y$ | 1 | 0 | $\theta_y \theta_{\bar{y}}$ |
| $\bar{x}\bar{y}z$ | 1 | $\theta_{\bar{y}}$ | $\theta_{\bar{y}}$ | $\theta_{z|xy}$ | $\theta_{\bar{y}} \theta_{z|xy}$ | $\theta_{\bar{y}} \theta_{\bar{y}} \theta_{z|xy}$ |

We have that $Pr(y|\mathbf{e}) = \theta_y$, $Pr(\mathbf{e}) = \theta_{z|xy}$ and that $Pr'(\mathbf{e}) \geq 2 \cdot \theta_y \theta_{\bar{y}}$. Since $\log x \leq x - 1$, we have

$$2 \cdot \theta_y \theta_{\bar{y}} = \theta_y \theta_{\bar{y}} + \theta_{\bar{y}} \theta_y$$
$$= \theta_y(1 - \theta_y) + \theta_{\bar{y}}(1 - \theta_{\bar{y}})$$
$$\leq -\theta_y \log \theta_y - \theta_{\bar{y}} \log \theta_{\bar{y}}$$
$$= ENT(Y|\mathbf{e}).$$

Thus, if we set $\theta_{z|xy} = (2 \cdot \theta_y \theta_{\bar{y}})^2$, then we get that:

$$\log \frac{Pr'(\mathbf{e})}{Pr(\mathbf{e})} \geq \log \frac{1}{2 \cdot \theta_y \theta_{\bar{y}}} \geq \log \frac{1}{ENT(Y|\mathbf{e})}.$$

So, as $\theta_y$, and thus $ENT(Y|\mathbf{e})$, goes to zero, we can set paramater $\theta_{z|xy}$ so that $\log Pr(\mathbf{e})/Pr'(\mathbf{e})$ goes to infinity.